%% file: main.tex
\pdfoutput=1

\documentclass[11pt]{article}
\usepackage{ACL2023}
\usepackage{times}
\usepackage{latexsym}
\usepackage{booktabs}
\usepackage{color}
\usepackage{xcolor,colortbl}
\usepackage[T1]{fontenc}
\usepackage[utf8]{inputenc}
\usepackage{microtype}
\usepackage{inconsolata}
\usepackage{graphicx}
\usepackage{enumitem}

\input{commands}
\title{Understand Legal Documents with Contextualized Large Language Models}

\author{Xin Jin$^*$ \\
  Northwest Polytechnical Unviersity \\
  \texttt{xin.jinsw@gmail.com} \\
  \And
  Yuchen Wang$^*$ \\
  Ohio State University\\
  \texttt{wang.9298@osu.edu} 
  }

\begin{document}

\maketitle

\def\thefootnote{*}\footnotetext{These authors contributed equally to this work.}\def\thefootnote{\arabic{footnote}}

\input{Section/abstract}

\input{Section/introduction}

\input{Section/background}

\input{Section/design}

\input{Section/evaluation}

\input{Section/conclusion}

\input{Section/acknowledge}

\bibliography{references}
\bibliographystyle{acl_natbib}

\appendix

\input{Section/appendix.tex}

\end{document}

%% file: Section/abstract.tex
\begin{abstract}

The growth of pending legal cases in populous countries, such as India, has become a major issue. Developing effective techniques to process and understand legal documents is extremely useful in resolving this problem. In this paper, we present our systems for SemEval-2023 Task 6: understanding legal texts~\cite{legaleval-2023}. Specifically, we first develop the Legal-BERT-HSLN model that considers the comprehensive context information in both intra- and inter-sentence levels to predict rhetorical roles (subtask A) and then train a Legal-LUKE model, which is legal-contextualized and entity-aware, to recognize legal entities (subtask B). Our evaluations demonstrate that our designed models are more accurate than baselines, e.g., with an up to 15.0\% better F1 score in subtask B. We achieved notable performance in the task leaderboard, e.g., 0.834 micro F1 score, and ranked No.5 out of 27 teams in subtask A.

\end{abstract}

%% file: Section/introduction.tex
\section{Introduction}

The growing amount of legal cases and documents requires more and more human efforts to process them.~\cite{kalamkar2022corpus,malik2021semantic}
In some countries, such as India, legal cases have accumulated in an incredible number. For example, India has more than 47 million cases pending in the courts~\cite{kalamkar2022corpus}. This has created a need for automated methods to help judges efficiently understand and process relevant and reliable information. In addition, these methods can also help students, legal scholars, and court officials who deal with legal documents daily. One way to assist them is to automatically understand and highlight the key information and context of long legal documents. 

However, understanding legal documents by machines is not an easy task. 
First, legal documents are often full of technical terminologies, which can span different legal divisions~\cite{kalamkar2022corpus}. 
Furthermore, legal documents can be specific to special cases, such as health~\cite{young2009legal}, IT technology~\cite{lu2017convolutional}, and cyber security~\cite{shackelford2016unpacking,jin2018multi,lu2017game}, which will involve domain-specific words.
In addition, legal documents can be extremely long~\cite{kalamkar2022corpus}, which makes dependency-based techniques and models~\cite{luo2016text}, such as RNN models, fail to extract the context information due to gradient vanishes.
Finally, typos and unstructured documents introduce noises~\cite{kalamkar2022corpus}, which makes automated natural language processing challenging.

Despite these challenges, predicting rhetorical roles and recognizing named entities in legal documents are very useful for automating the processing and understanding of legal documents. Rhetorical role prediction segments texts and structures noisy legal documents into topically and semantically coherent units~\cite{ghosh2019identification}. Named entity recognition helps identify key legal entities in long documents~\cite{nadeau2007survey}, which can not only help judges process cases in a more efficient way, but also benefit the next automation steps. These two tasks can serve as key steps in these methods~\cite{legaleval-2023}.

In this paper, we propose to solve the rhetorical role prediction and named entity recognition problems in the legal document domain with contextualized large-language models. 
We first systematically build models with well-known design choices based on popular pre-trained models (e.g., BERT and XLM-roBERTa)~\cite{qiu2020pre}, then systematically evaluate the performance of different models and identify the key limitations, and eventually propose our legal contextualized models as our final solutions.
For rhetorical role prediction, we model this task as the sequential sentence classification problem and build the Legal-BERT-HSLN model, which considers the comprehensive context semantics in both intra- and inter-sentence levels.
For named entity recognition, we propose to build a legal-LUKE model that is both sensitive to context and entity-aware.
Our evaluation results show that our proposed models are more accurate than baselines, e.g., Legal-LUKE is 15.0\% better than our baseline BERT-CRF in F1 score (more details in \S\ref{sec:ner-results}). Furthermore, we also achieved the top performance on the rhetorical role prediction task leaderboard, i.e., ranked No.5  out of 27 teams and achieved the 0.8343 micro F1 score (see \S\ref{sec:rr-results}).

We briefly summarize
our primary contributions as follows.
\begin{itemize}[noitemsep]
\item We formalize the rhetorical role prediction task as a sequential sentence classification problem and build the Legal-BERT-HSLN framework to model comprehensive sentence semantics. 
\item We construct the legal-LUKE model with contextualized legal-document and entity-aware representations.
\item Our evaluations demonstrate the better performance of our proposed model compared to baselines and achieved promising results on the task leaderboard.
\end{itemize}

%% file: Section/background.tex
\section{Background}

\subsection{Sequential Sentence Classification}

Sequential sentence classification is a natural language processing (NLP) technique that involves classifying a sequence of sentences into one or more categories or labels~\cite{hassan2017deep,cohan2019pretrained,jin2018hierarchical,brack2022cross}. 
The objective of this technique is to analyze and classify the content of a given text document based on the semantics and context of the sentences~\cite{jin2022understanding}.
It is often used in classification tasks~\cite{xu2018end,shan2021nrtsi,shan2022transparent,baskaran2022distribution}, such as sentiment analysis, spam detection, and topic classification, where the classification of a single sentence can depend on the preceding or succeeding sentences~\cite{hassan2017deep}. This technique can be implemented using various algorithms, such as recurrent neural networks (RNNs) or long-short-term memory (LSTM) networks, which are capable of processing sequential data~\cite{lipton2015critical, luo2017text}.

In sequential sentence classification, the input is a sequence of sentences and the output is one or more labels that describe the content of the document. The classification can be performed at the sentence level or at the document level, depending on the specific use case~\cite{cohan2019pretrained}. Sequential sentence classification is an important technique in NLP because it enables machines to understand and analyze the meaning and context of human language, which is crucial for many applications such as automated text summarization, QA agents and recommendation systems ~\cite{qiu2020pre,li2020sampling,jin2021estimating,liu2020large}.

\subsection{Legal Named Entity Recognition}

The objective of named entity recognition in the legal domain is to detect and label all instances of specific legally relevant named entities within unstructured legal reports~\cite{nadeau2007survey,kalamkar2022named,li2020survey}. 
Using this named entity information, one can analyze, aggregate, and mine data to uncover insightful patterns. 
Furthermore, the ultimate goal of legal document analysis is to automate the process of information retrieval or mapping a legal document to one or more nodes of a hierarchical taxonomy or legal cases, in which legal NER plays a significant role.~\looseness=-1 

Given that legal reports contain a large number of complex medical terms and terminologies, such as statute and precedent, identifying expressions referring to anatomies, findings, and anatomical locations is a crucial aspect in organizing information and knowledge within the legal domain~\cite{kalamkar2022named}. 
Therefore, automating this identification process has been recognized as a key target for automation. 
Moreover, automatic named entity recognition (NER) helps exhaustively extract semantic information and misspelling checking~\cite{kalamkar2022named,legaleval-2023}.

%% file: Section/design.tex
\section{System Overview}

	
	
	
	
	

\subsection{Rhetorical Role Classification}

Identifying rhetorical roles (RR) in legal documents is a challenging task, which requires machine learning models to accurately classify sentences into predefined RR categories. One of the primary challenges is the variability and complexity of natural language, including variations in sentence structure, word choice, and context. This requires training machine learning models on diverse and large datasets that capture the range of languages used in the real world. 
Another significant challenge is capturing the long-term dependencies between sentences in a sequence, which can be difficult to achieve but is essential to determine the overall meaning and context of the text. 

To solve this problem, we first observe that this task is basically a sequential sentence classification problem. Meanwhile, existing work proposes to solve this problem using encoder-decoder models, where the encoder embeds the sentence-level semantics and considers the context dependency information, and the decoder classifies individual sentences by the contextualized surrounding sentence information. For example, Jin et. al~\cite{jin2018hierarchical} classify medical abstract sentences with a hierarchical sequential labeling network, which is composed of a word encoder, a sentence encoder, and a document encoder. Specifically, the word-level encoder is a bidirectional LSTM that encodes each word in a sentence into a vector~\cite{zhang11fusion}. The sentence-level encoder is another bidirectional LSTM that encodes each sentence vector in a hidden state. The document-level encoder is either an LSTM or a CRF layer that models the dependencies between sentences and outputs a label for each sentence. 

\begin{figure}
\centering
\includegraphics[width=0.48\textwidth]{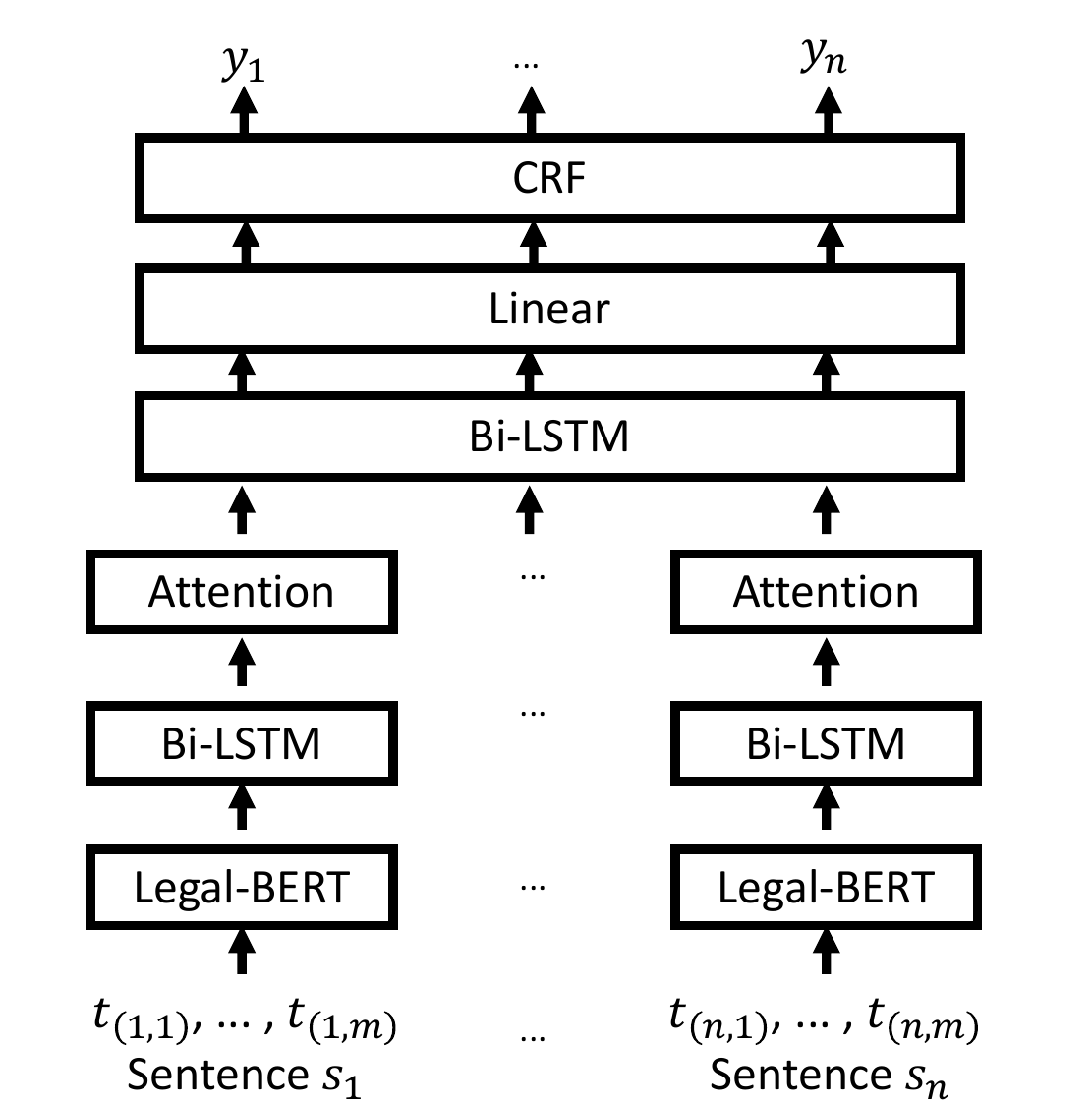}
\caption{Model Architecture of Legal-BERT-HSLN}
\label{fig:Legal-BERT-HSLN}
\vspace{-0.1in}
\end{figure}

We follow the paradigm and propose the Legal-BERT-HSLN model for legal rhetorical role classification. In our design, we applied the HSLN structure from Brack et al.~\cite{brack2022cross} and changed the backbone model to Legal-BERT~\cite{chalkidis2020legal}. The model architecture of Legal-BERT-HSLN is shown in figure~\autoref{fig:Legal-BERT-HSLN}. 
The model first takes as input the sequences of legal word tokens (\{$t_{i,1}, t_{i,2}, ..., t_{i,m}$\}) of sentence $s_{i}$ and Legal-BERT generates the corresponding token embeddings. Next, the token embeddings are further enriched with local context information within the sentence $s_{i}$ by Bi-LSTM and the attention pooling layer as the augmented token embeddings (\{$e_{i,1}, e_{i,2}, ..., e_{i,m}$\}), which is aggregated to generate the embedding of the sentence $e_{s_{i}}$. Since one of the most important features in this task is the inter-sentence dependency, we further enrich the sentence embedding with contextual information from surrounding sentences. The output layer transforms contextualized sentence embeddings into rhetorical role labels via a linear transformation and CRF. We also introduce dropout layers after each layer for regularization.

\subsection{Legal Named Entity Recognition}

In SemEval task 6~\cite{legaleval-2023}, the second subtask is the recognition of legal entities named entities. 
Specifically, the legal documents provide nonexhaustive metadata with 14 legal entities, including petitioner, respondent, court, statute, provision, precedents, etc. 
Identifying these legal entities can be both error prone and labor intensive~\cite{mohit2014named}. 

Although the task of legal NER may seem straightforward, it is in fact a challenging undertaking due to several reasons. 
First, the NER task itself is an unsolved problem~\cite{li2020survey}. 
Moreover, legal documents can be very noisy~\cite{kalamkar2022corpus,kalamkar2022named}. 
Legal texts contain morphological forms (e.g., synonyms, abbreviations, and even misspellings), which means that different legal cases can use different words and phrases to express the same meaning. 
To process the natural language texts, existing approaches use vocabulary-based embeddings~\cite{wang2020static}. 
However, legal documents involve many out-of-vocabulary words~\cite{jin2022symlm}, which further makes the task complex. 

Meanwhile, we have several insights to solve the problems. First, natural language preprocessing, e.g., tokenization, POS tagging, and sentence parsing, can mitigate the noise of legal documents. 
Moreover, the identified POS tags and sentence structure can help determine the legal entities by the nature of human languages. 
Second, compared to static entity representations that assign fixed embeddings to words and entities in the knowledge base, contextualized word representations can generate adaptive semantic embeddings of words and entities, which can be tuned by domain-specific context~\cite{ethayarajh2019contextual}. 
Finally, we observe that state-of-the-art entity-aware representations~\cite{yamada-etal-2020-luke,ri-etal-2022-mluke} can take advantage of both the advantages of contextualized word embeddings and create entity representations based on the rich entity-centric semantics encoded in the corresponding entity embeddings.
Therefore, we combine these insights and propose to identify legal entities with embeddings of legal-sensitive entities.

\begin{figure}
\centering
\includegraphics[width=0.48\textwidth]{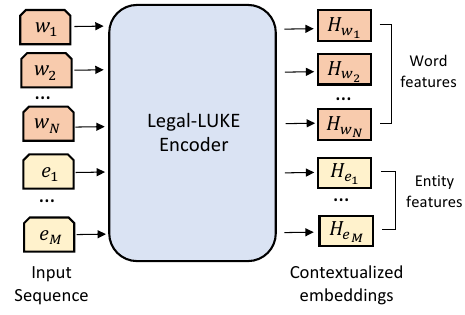}
\caption{Entity-aware Contextualized Representation}
\label{fig:legal-luke}
\vspace{-0.1in}
\end{figure}

For legal NER, we introduce a legal entity recognition model: Legal-LUKE, based on the bidirectional transformer encoder of LUKE~\cite{yamada-etal-2020-luke}. The LUKE model was pre-trained by predicting both words and entities masked by the \texttt{[MASK]} tokens. As shown in~\autoref{fig:legal-luke}, Legal-LUKE takes as input the sequence of preprocessed words ($w_1, w_2, .., w_N$) and entities ($e_1, e_2, ..., e_M$). The encoder generates the legal-contextualized representations for both words and entities, i.e., ($H_{w_1}, H_{w_2}, .., H_{w_N}$) and ($H_{e_1}, H_{e_2}, .., H_{e_M}$). To compute the word and entity embeddings, three embeddings are added together, which include token embeddings, type embeddings, and position embeddings. Position embeddings are used to associate entity tokens with their corresponding word tokens, where the position of an entity token is determined by the positions of its corresponding word tokens. The entity position embeddings are then added up over these positions.

%% file: Section/evaluation.tex
\section{Evaluations}

We conduct extensive experiments using the rhetorical role classification and legal named entity recognition tasks. 
We have implemented the Legal-BEST-HSLN model and the Legal-LUKE model based on Pytorch~\cite{paszke2019pytorch}. 

\subsection{Experimental Setup}

The input word sequence is created by inserting the tokens of \texttt{[CLS]} and \texttt{[SEP]} into the original word sequence as the first and last tokens, respectively, unless stated otherwise.
For the input entity sequence of Legal-LUKE, the legal entities from the training set are used to fine-tune the model.
Although we cannot access the test set before the task organizers open the evaluation system, we train and tune our models and baselines based on the training and validation set.

\subsubsection{Baselines for Rhetorical Role Classification}

For subtask A, we set up our experiments by selecting BERT-base and 3 BERT variants with minor modifications as our baselines:
\begin{itemize}[noitemsep]
    \item \textbf{BERT-Base}. For this model, we directly use the \texttt{[CLS]} token embedding as the sentence embedding as the encoder of the classifier.
    \item \textbf{BERT-Mean}. Instead of using the hidden state of the token \texttt{[CLS]} for the classification output from BERT, we tried to use the mean value of 128 lengths of sequences.
    \item \textbf{BERT-Regularization}. In this method, we made a preprocessing work to the training set by regularizing symbols. These procedures include: (a) lowercase, (b) remove @name, (c) isolate and remove punctuation except ‘?’, (d) remove special characters such as columns and semi-columns, (e) remove stop words except ‘not’ and ‘can’, and (f) remove trailing whitespace.
    \item \textbf{BERT-Augmentation}. Each sentence training set is randomly swapped and the entire training size is doubled.
\end{itemize}

\subsubsection{Baselines for Legal Named Entity Recognition}

For subtask B, we select the following 3 models as our baselines:

\begin{itemize}[noitemsep]
    \item \textbf{BERT-CRF}. For this model, we use the BERT model as the encoder and transform the encodings into NER labels with the CRF prediction head.
    \item \textbf{BERT-Span}. BERT-Span~\cite{joshi2020spanbert} is the variation of BERT that uses BERT to train span boundary representations, which are more suitable for entity boundary detection.
    \item \textbf{XLM-roBERTa-CRF}. XLM-roBERTa-CRF  is a combination of XLM-RoBERTa~\cite{conneau2019unsupervised} and CRF. XLM-RoBERTa is a multilingual version of RoBERTa, a transformer model pre-trained on a large corpus in a self-supervised fashion. We used XLM-RoBERTa-large as the encoder model.
    \item \textbf{mLUKE}. mLUKE~\cite{ri2021mluke} is a multilingual extension of LUKE, a pre-trained language model that incorporates entity information from Wikipedia.
\end{itemize}


\subsubsection{Test Environment} The evaluations are performed on a desktop server, with the Intel Xeon E5-1650 CPU, Ubuntu 18.04 OS, 64 GB memory, 4 TB storage, and 4 NVIDIA GeForce GTX 1080 Ti graphics cards. While training the Legal-LUKE model, we encountered out-of-memory issues, and we switched to another server, which has the Intel Xeon W-2245 CPU, Ubuntu 20.04 OS, 128 GB memory, 2 TB storage, and an NVIDIA RTX A5000 graphics card.


\subsection{Results of Rhetorical Role Classification}
\label{sec:rr-results}

The goal of rhetorical role classification is to provide a categorization label for each sentence at the unstructured document level. The number of classes is 13. The training set has 247 documents with a total sentence number of ~30k, validation set has 30 documents with a total sentence number of ~3k. In this section, all evaluations are reported based on the validation set.

\begin{table}[]
\centering
\resizebox{0.45\textwidth}{!}{
\begin{tabular}{lrr}
\toprule
\textbf{Model}               & \textbf{Micro F1 Score} & \textbf{\textbf{Best Epoch}} \\
\midrule
BERT-Base           & 0.631                              & 5                              \\
BERT-Mean           & 0.641                              & 4                              \\
BERT-Regularization & 0.597                              & 4                              \\
BERT-Augmentation   & 0.645                              & 4                             \\

\textbf{Legal-BERT-HSLN}     & \textbf{0.828}                             & \textbf{16}
\\
\bottomrule
\end{tabular}
}
\caption{Rhetorical Role Classification Performance}
\label{tab:rr-summary}
\end{table}

\autoref{tab:rr-summary} shows the summary of micro F1 scores for Legal-BERT-HSLN and the baselines on the validation set, where Legal-BERT-HSLN significantly outperforms all baselines and it takes more epochs to converge, where we believe Legal-BERT-HSLN has the better capacity to understand legal documents. It is worth mentioning that the regularization process decreased the F1 score and suggests that the rhetorical role is sensitive to detailed stop words and external sentence markers.

\begin{figure}
\centering
\includegraphics[width=0.48\textwidth]{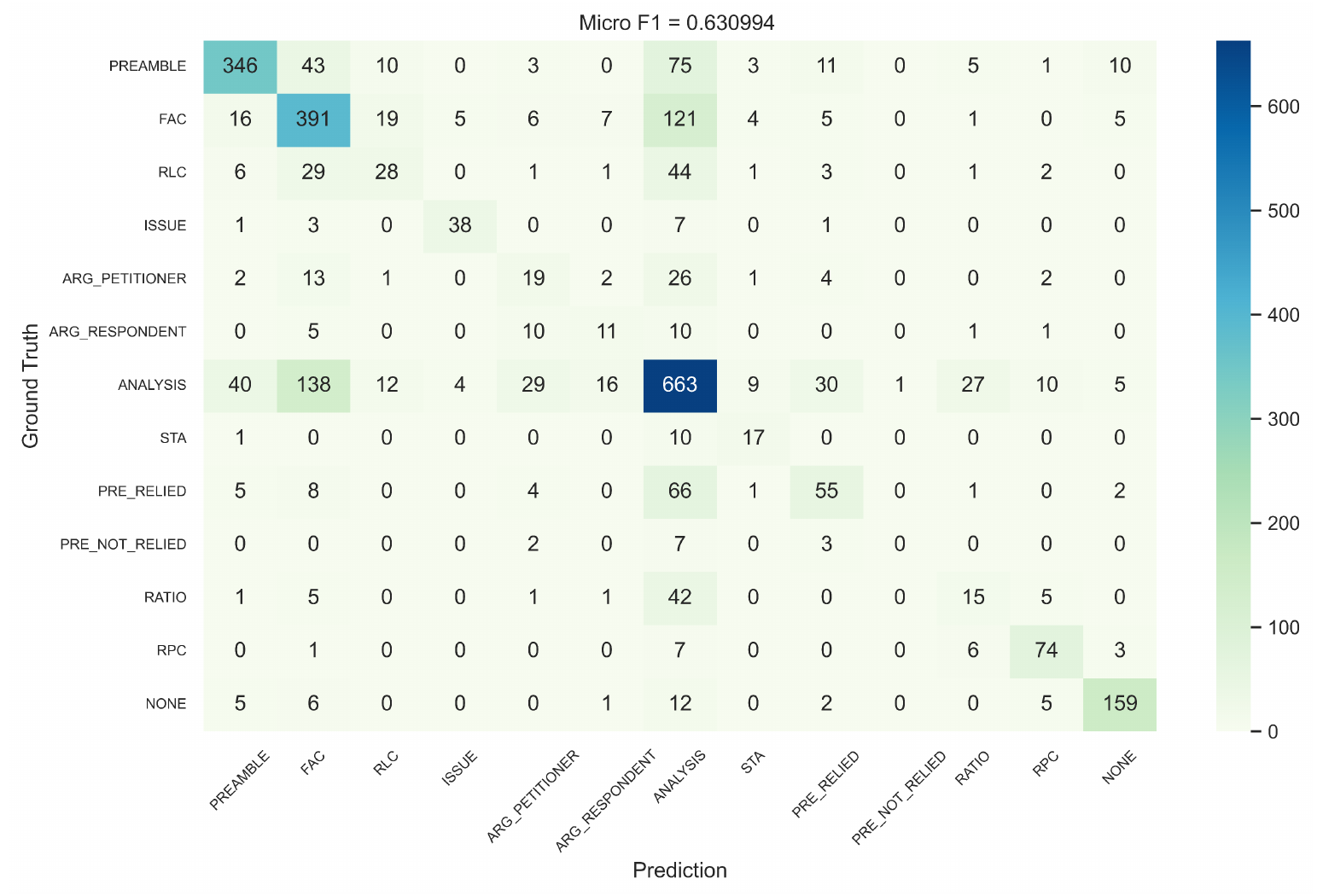}
\caption{Confusion Matrix of BERT Baseline Predictions}
\label{fig:baseline-confusion}
\end{figure}

While our initial design choice was not Legal-BERT-HSLN, we originally trained the baseline models. 
Specifically, our first design choice was to treat all sentences as individual elements where they have no correlation to each other. We set the baseline model with the BERT network backbone with a simple multilayer perceptron. The resulting confusion matrix is shown in \autoref{fig:baseline-confusion} with the micro F1 0.631. 
Subsequently, we conducted 3 BERT variants with minor modifications and obtained the performance shown in \autoref{tab:rr-summary}. From the results, we observed the significant performance gap between all the BERT baselines and the state-of-the-art solution~\cite{kalamkar2022corpus}.

\begin{figure}
\centering
\includegraphics[width=0.4\textwidth]{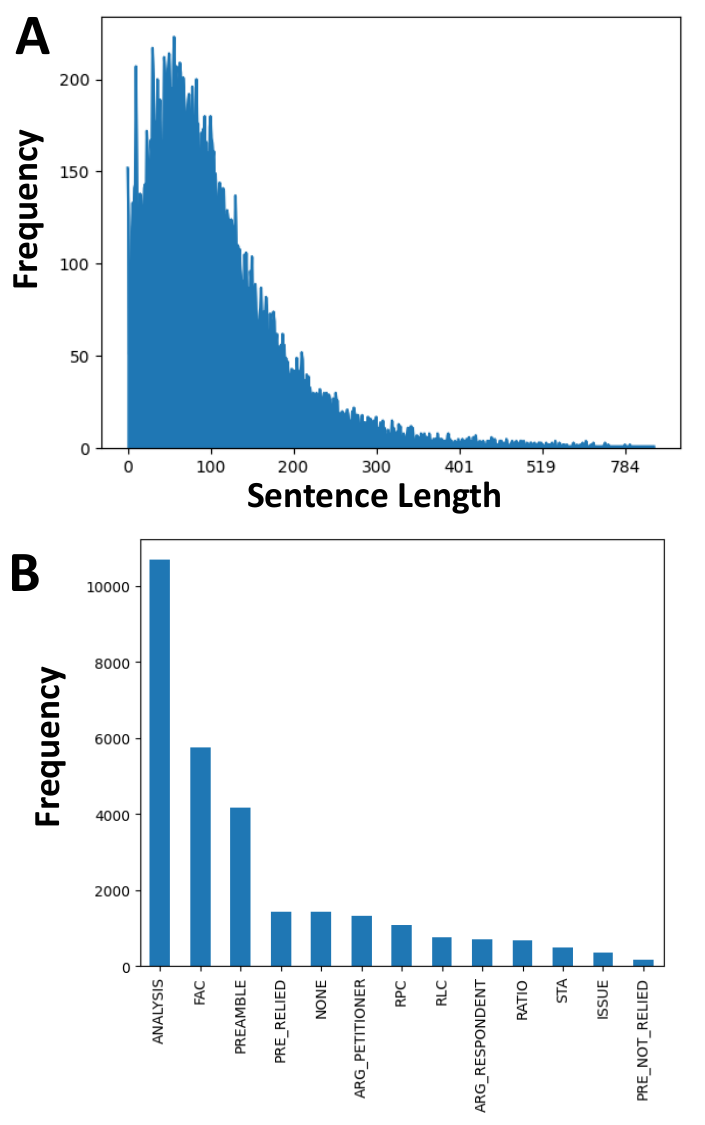}
\caption{(A) Sentence Length and (B) Class Distributions in Rhetorical Role Dataset}
\label{fig:RR-dataset}
\vspace{-0.1in}
\end{figure}

To fully understand the problem, we put an effort into examining the distribution of the legal dataset~\cite{legaleval-2023} in term of sentence length and classes.
In~\autoref{fig:RR-dataset} (A), the sentence lengths overall satisfied Poisson’s distribution but there is a significant noise at <20 levels. These short sentences may influence the accuracy of classification due to a lack of meaningful information. For example, sentences 17928, 17929, and 17930 are "\textit{[326C-E}", "\textit{] 2.}", and "\textit{There may be circumstances where expenditure, even if incurred for obtaining advantage of enduring benefit would not amount to acquisition of asset}", respectively, whose ground-truth labels are all PREMABLE. This might be due to the discontinuous annotation from turning the page or a keyboard input mistake. Therefore, model prediction from the sentence level is likely not an optimal paradigm to proceed. In~\autoref{fig:RR-dataset} (B), we found that there is a significant bias of the labeling in the training set; therefore, we believe that more experiments, such as validating the label bias, can be tested for future research.

\begin{figure}
\centering
\includegraphics[width=0.48\textwidth]{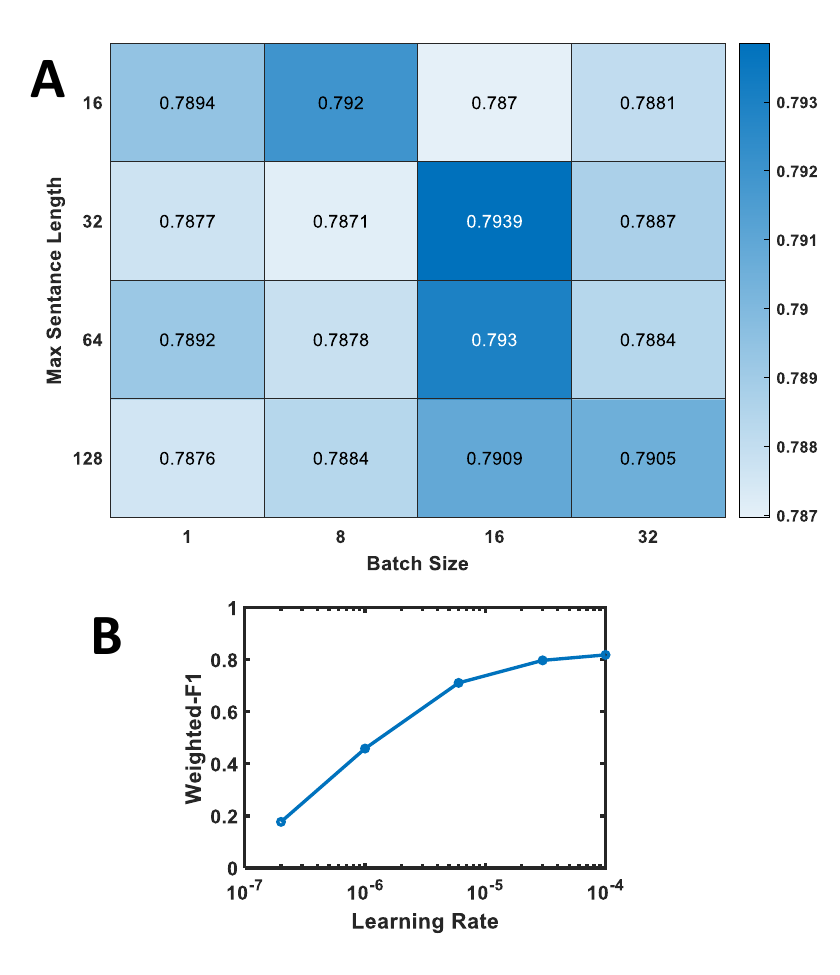}
\caption{Legal-BERT-HSLN Performance Affected by (A) Maximum Sentence Length and Batch Size and (B) Learning Rate}
\label{fig:hyper-param}
\end{figure}

\begin{figure}
\centering
\includegraphics[width=0.48\textwidth]{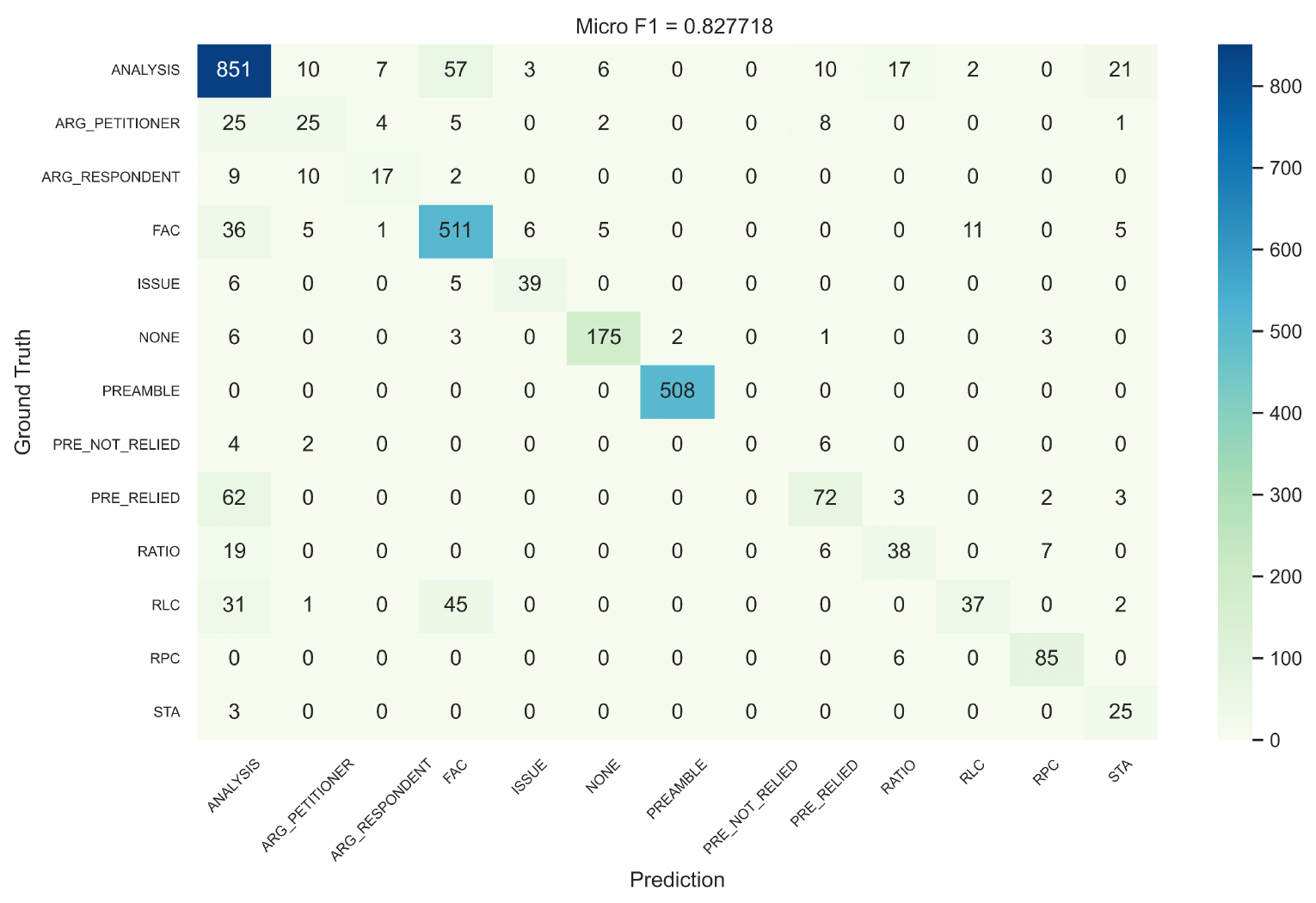}
\caption{Confusion Matrix with Legal-BERT-HSLN}
\label{fig:Legal-BERT-HSLN-confusion}
\end{figure}

Except for the design choices, we also evaluate the effectiveness of our hyper-parameters. 
\autoref{fig:hyper-param} shows the micro F1 score is influenced by the batch size, maximum sentence length, and learning rate, where the best combination we have tested so far is batch size 16, maximum sentence length 32, learning rate $1\times10^{-4}$.
All weighted-F1 scores in \autoref{fig:hyper-param} are calculated from \texttt{sklearn.metrics.f1\_score} function with the \texttt{weighted} option. 
\autoref{fig:Legal-BERT-HSLN-confusion} shows the confusion matrix using this combination of hyperparameters, and it is found that the best Micro F1 score we achieve in the validation set is 0.828.
Although the major error contribution is no longer in the PREMABLE label, the accuracy of the PRE\_NOT\_RELIED label is 0 which is the same as reported by the state-of-the-art solution~\cite{kalamkar2022corpus}. For future experiments, we suggest studying this outlier effect.

\paragraph{Performance on Test Set} After the task organizer released the test, we test Legal-BEST-HSLN and submitted the predictions. Surprisingly, we obtained the micro F1 score of 0.8343 and ranked No. 5 out of 27 teams. Note that the test performance is better than all performance on the validation set. Therefore, we believe that there is a shift in the distribution of the training, validation, and test sets.

\subsection{Results of Legal Named Entity Recognition}
\label{sec:ner-results}

\begin{table}[]
\centering
\resizebox{0.45\textwidth}{!}{
\begin{tabular}{lrr}
\toprule
\textbf{Model}               & \textbf{F1 Score} & \textbf{\textbf{Best Epoch}} \\
\midrule
BERT-CRF           & 0.694                              & 12                              \\
BERT-Span           & 0.712                              & 14                              \\ 
XLM-roBERTa-CRF & 0.773                       & 21                              \\
mLUKE   & 0.787                              & 12                             \\
\textbf{Legal-LUKE}     & \textbf{0.796}      & \textbf{18}
\\
\bottomrule
\end{tabular}
}
\caption{Overall Performance on Validation Set of Legal Named Entity
Recognition}
\label{tab:ner-summary}
\end{table}

\autoref{tab:ner-summary} presents the overall performance of Legal-LUKE and the baseline models for legal named entity recognition on the validation set.
We observe that Legal-LUKE outperforms all baselines with an up to 14.3\% better micro F1 score. 

\begin{figure}
\centering
\includegraphics[width=0.4\textwidth]{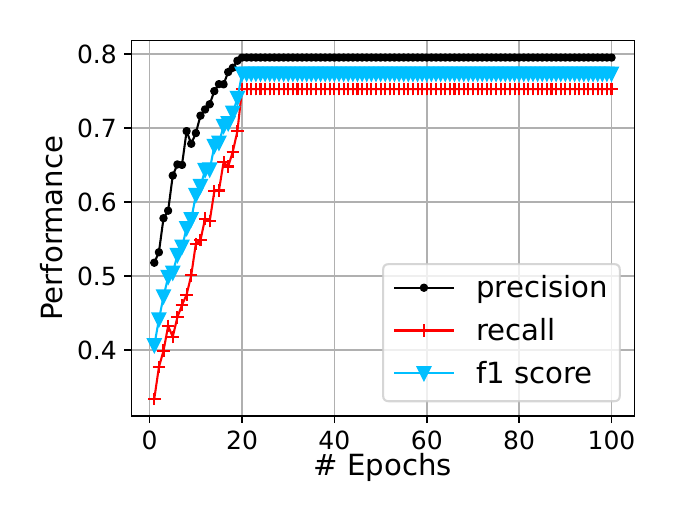}
\vspace{-0.2in}
\caption{The Best Performance of the Baseline Model XLM-roBERTa-CRF across Training Epochs}
\label{fig:ner-baseline}
\end{figure}

Moreover, among the baselines, the BERT-Span model is better than the BERT-CRF model where we used the same encoder model but different prediction heads, which demonstrates the benefits of using the span boundary representations for the named entity recognition task. Moreover, the XLM-roBERTa encoder model shows better legal text encoding capacity because the XLM-roBERTa-CRF model and BERT-CRF model share the same decoder module, but XLM-roBERTa-CRF outperforms BERT-CRF with a 11.4\% better F1 score. Note that we gave the same parameters for the baseline models. For example, we consider context information by a natural language preprocessor with a maximum length of 100. To balance train efficiency and performance, we set the batch size per GPU as 16 with a learning rate scheduler whose initial rate is $5.0\times 10^{-6}$ and the AdamW optimizer. Although we set the maximum number of epochs as 10, we observed that all models achieved the best performance at early epochs. For example, \autoref{fig:ner-baseline} shows the best performance of the XLM-roBERTa-CRF baseline model on the validation set in different epochs, which shows the convergence effect as the training epoch increases.

\begin{figure}
\centering
\includegraphics[width=0.4\textwidth]{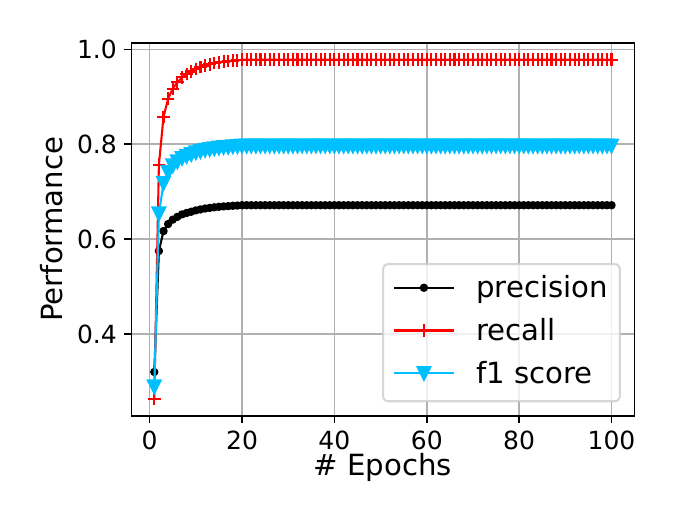}
\vspace{-0.2in}
\caption{The Best Performance of Legal-LUKE across Training Epochs}
\label{fig:ner-legal-luke}
\end{figure}

For our Legal-LUKE model, we also use a learning rate scheduler that has a warm-up ratio of 0.06, the same initial rate $5.0\times 10^{-6}$ as baselines, and the AdamW optimizer. While the pre-trained LUKE model is too large for our GPU server, we tune the different batch sizes to trade off between training speed and performance, and eventually we set the batch size as 8. Similar to the baseline models, we also observe the performance convergence (as shown in~\autoref{fig:ner-legal-luke}) though we set the maximum epoch number at 100.

\paragraph{Performance on Test Set} After the task organizer released the test, we evaluated Legal-LUKE on the test set and submitted the predictions. We obtained an F1 score of 0.667 and ranked No. 13 out of 17 teams. Our test F1 score is far lower than the validation F1 score. However, according to the dataset~\cite{kalamkar2022named}, the validation set and test set should have similar distributions as they are from the same group of legal cases. Our hypothesis is that the event organizer adopted a different preprocessor from ours which produces different token indices and label shifts. For example, we observed that the legal documents contain many empty lines and we opted to remove such lines in our processing, which can result in the different token indices in our predictions. We plan to further investigate the root cause upon the release of the test set.

%% file: Section/conclusion.tex
\section{Conclusion}
In this paper, we propose to understand legal documents with context-sensitive language models. For  the sub-task of rhetorical role classification, we design the Legal-BERT-HSLN model, which learns the hierarchical context information to solve the sequential sentence classification problem. For legal named entity recognition, we implemented the Legal-LUKE model which is both contextualized and entity-aware. Our evaluation results reveal the outperformance of our models compared to baselines and we are the top-5 teams for the  rhetorical role classification task on the leaderboard.

%% file: Section/acknowledge.tex

%% file: Section/appendix.tex


%% file: main.bbl
\begin{thebibliography}{40}
\expandafter\ifx\csname natexlab\endcsname\relax\def\natexlab#1{#1}\fi

\bibitem[{Baskaran et~al.(2022)Baskaran, Ranek, Shan, Stanley, and
  Oliva}]{baskaran2022distribution}
Vishal~Athreya Baskaran, Jolene Ranek, Siyuan Shan, Natalie Stanley, and
  Junier~B Oliva. 2022.
\newblock Distribution-based sketching of single-cell samples.
\newblock In \emph{Proceedings of the 13th ACM International Conference on
  Bioinformatics, Computational Biology and Health Informatics}, pages 1--10.

\bibitem[{Brack et~al.(2022)Brack, Hoppe, Buscherm{\"o}hle, and
  Ewerth}]{brack2022cross}
Arthur Brack, Anett Hoppe, Pascal Buscherm{\"o}hle, and Ralph Ewerth. 2022.
\newblock Cross-domain multi-task learning for sequential sentence
  classification in research papers.
\newblock In \emph{Proceedings of the 22nd ACM/IEEE Joint Conference on Digital
  Libraries}, pages 1--13.

\bibitem[{Chalkidis et~al.(2020)Chalkidis, Fergadiotis, Malakasiotis, Aletras,
  and Androutsopoulos}]{chalkidis2020legal}
Ilias Chalkidis, Manos Fergadiotis, Prodromos Malakasiotis, Nikolaos Aletras,
  and Ion Androutsopoulos. 2020.
\newblock Legal-bert: The muppets straight out of law school.
\newblock \emph{arXiv preprint arXiv:2010.02559}.

\bibitem[{Cohan et~al.(2019)Cohan, Beltagy, King, Dalvi, and
  Weld}]{cohan2019pretrained}
Arman Cohan, Iz~Beltagy, Daniel King, Bhavana Dalvi, and Daniel~S Weld. 2019.
\newblock Pretrained language models for sequential sentence classification.
\newblock \emph{arXiv preprint arXiv:1909.04054}.

\bibitem[{Conneau et~al.(2019)Conneau, Khandelwal, Goyal, Chaudhary, Wenzek,
  Guzm{\'a}n, Grave, Ott, Zettlemoyer, and Stoyanov}]{conneau2019unsupervised}
Alexis Conneau, Kartikay Khandelwal, Naman Goyal, Vishrav Chaudhary, Guillaume
  Wenzek, Francisco Guzm{\'a}n, Edouard Grave, Myle Ott, Luke Zettlemoyer, and
  Veselin Stoyanov. 2019.
\newblock Unsupervised cross-lingual representation learning at scale.
\newblock \emph{arXiv preprint arXiv:1911.02116}.

\bibitem[{Ethayarajh(2019)}]{ethayarajh2019contextual}
Kawin Ethayarajh. 2019.
\newblock How contextual are contextualized word representations? comparing the
  geometry of bert, elmo, and gpt-2 embeddings.
\newblock \emph{arXiv preprint arXiv:1909.00512}.

\bibitem[{Ghosh and Wyner(2019)}]{ghosh2019identification}
Saptarshi Ghosh and Adam Wyner. 2019.
\newblock Identification of rhetorical roles of sentences in indian legal
  judgments.
\newblock In \emph{Legal Knowledge and Information Systems: JURIX 2019: The
  Thirty-second Annual Conference}, volume 322, page~3. IOS Press.

\bibitem[{Hassan and Mahmood(2017)}]{hassan2017deep}
Abdalraouf Hassan and Ausif Mahmood. 2017.
\newblock Deep learning for sentence classification.
\newblock In \emph{2017 IEEE Long Island Systems, Applications and Technology
  Conference (LISAT)}, pages 1--5. IEEE.

\bibitem[{Jin and Szolovits(2018)}]{jin2018hierarchical}
Di~Jin and Peter Szolovits. 2018.
\newblock Hierarchical neural networks for sequential sentence classification
  in medical scientific abstracts.
\newblock \emph{arXiv preprint arXiv:1808.06161}.

\bibitem[{Jin et~al.(2021)Jin, Li, Mudrak, Gao, and Liu}]{jin2021estimating}
Ruoming Jin, Dong Li, Benjamin Mudrak, Jing Gao, and Zhi Liu. 2021.
\newblock On estimating recommendation evaluation metrics under sampling.
\newblock In \emph{Proceedings of the AAAI Conference on Artificial
  Intelligence}, volume~35, pages 4147--4154.

\bibitem[{Jin et~al.(2018)Jin, Lu, Liu, and Zhu}]{jin2018multi}
Xin Jin, Wei Lu, Siqi Liu, and Zuqing Zhu. 2018.
\newblock On multi-layer restoration in optical networks with encryption
  solution deployment.
\newblock In \emph{2018 Optical Fiber Communications Conference and Exposition
  (OFC)}, pages 1--3. IEEE.

\bibitem[{Jin et~al.(2022{\natexlab{a}})Jin, Manandhar, Kafle, Lin, and
  Nadkarni}]{jin2022understanding}
Xin Jin, Sunil Manandhar, Kaushal Kafle, Zhiqiang Lin, and Adwait Nadkarni.
  2022{\natexlab{a}}.
\newblock Understanding iot security from a market-scale perspective.
\newblock In \emph{Proceedings of the 2022 ACM SIGSAC Conference on Computer
  and Communications Security}, pages 1615--1629.

\bibitem[{Jin et~al.(2022{\natexlab{b}})Jin, Pei, Won, and Lin}]{jin2022symlm}
Xin Jin, Kexin Pei, Jun~Yeon Won, and Zhiqiang Lin. 2022{\natexlab{b}}.
\newblock Symlm: Predicting function names in stripped binaries via
  context-sensitive execution-aware code embeddings.
\newblock In \emph{Proceedings of the 2022 ACM SIGSAC Conference on Computer
  and Communications Security}, pages 1631--1645.

\bibitem[{Joshi et~al.(2020)Joshi, Chen, Liu, Weld, Zettlemoyer, and
  Levy}]{joshi2020spanbert}
Mandar Joshi, Danqi Chen, Yinhan Liu, Daniel~S Weld, Luke Zettlemoyer, and Omer
  Levy. 2020.
\newblock Spanbert: Improving pre-training by representing and predicting
  spans.
\newblock \emph{Transactions of the Association for Computational Linguistics},
  8:64--77.

\bibitem[{Kalamkar et~al.(2022{\natexlab{a}})Kalamkar, Agarwal, Tiwari, Gupta,
  Karn, and Raghavan}]{kalamkar2022named}
Prathamesh Kalamkar, Astha Agarwal, Aman Tiwari, Smita Gupta, Saurabh Karn, and
  Vivek Raghavan. 2022{\natexlab{a}}.
\newblock Named entity recognition in indian court judgments.
\newblock \emph{arXiv preprint arXiv:2211.03442}.

\bibitem[{Kalamkar et~al.(2022{\natexlab{b}})Kalamkar, Tiwari, Agarwal, Karn,
  Gupta, Raghavan, and Modi}]{kalamkar2022corpus}
Prathamesh Kalamkar, Aman Tiwari, Astha Agarwal, Saurabh Karn, Smita Gupta,
  Vivek Raghavan, and Ashutosh Modi. 2022{\natexlab{b}}.
\newblock Corpus for automatic structuring of legal documents.
\newblock \emph{arXiv preprint arXiv:2201.13125}.

\bibitem[{Li et~al.(2020{\natexlab{a}})Li, Jin, Gao, and Liu}]{li2020sampling}
Dong Li, Ruoming Jin, Jing Gao, and Zhi Liu. 2020{\natexlab{a}}.
\newblock On sampling top-k recommendation evaluation.
\newblock In \emph{Proceedings of the 26th ACM SIGKDD International Conference
  on Knowledge Discovery \& Data Mining}, pages 2114--2124.

\bibitem[{Li et~al.(2020{\natexlab{b}})Li, Sun, Han, and Li}]{li2020survey}
Jing Li, Aixin Sun, Jianglei Han, and Chenliang Li. 2020{\natexlab{b}}.
\newblock A survey on deep learning for named entity recognition.
\newblock \emph{IEEE Transactions on Knowledge and Data Engineering},
  34(1):50--70.

\bibitem[{Lipton et~al.(2015)Lipton, Berkowitz, and Elkan}]{lipton2015critical}
Zachary~C Lipton, John Berkowitz, and Charles Elkan. 2015.
\newblock A critical review of recurrent neural networks for sequence learning.
\newblock \emph{arXiv preprint arXiv:1506.00019}.

\bibitem[{Liu et~al.(2020)Liu, Huang, Gao, Chen, and Li}]{liu2020large}
Zhi Liu, Yan Huang, Jing Gao, Li~Chen, and Dong Li. 2020.
\newblock Large-scale real-time personalized similar product recommendations.
\newblock \emph{arXiv preprint arXiv:2004.05716}.

\bibitem[{Lu et~al.(2017{\natexlab{a}})Lu, Jin, and Zhu}]{lu2017game}
Wei Lu, Xin Jin, and Zuqing Zhu. 2017{\natexlab{a}}.
\newblock Game theoretical flexible service provisioning in ip over elastic
  optical networks.
\newblock In \emph{2017 Opto-Electronics and Communications Conference (OECC)
  and Photonics Global Conference (PGC)}, pages 1--3. IEEE.

\bibitem[{Lu et~al.(2017{\natexlab{b}})Lu, Zhang, Zhou, Liu, Jin, Guo, and
  Cao}]{lu2017convolutional}
Xiaofeng Lu, Ruonan Zhang, Yuliang Zhou, Jiawei Liu, Xin Jin, Qi~Guo, and Chang
  Cao. 2017{\natexlab{b}}.
\newblock Convolutional modeling and antenna de-embedding for wideband spatial
  mmwave channel measurement.
\newblock In \emph{2017 IEEE Wireless Communications and Networking Conference
  (WCNC)}, pages 1--6. IEEE.

\bibitem[{Luo and Huang(2017)}]{luo2017text}
Yubo Luo and Yongfeng Huang. 2017.
\newblock Text steganography with high embedding rate: Using recurrent neural
  networks to generate chinese classic poetry.
\newblock In \emph{Proceedings of the 5th ACM workshop on information hiding
  and multimedia security}, pages 99--104.

\bibitem[{Luo et~al.(2016)Luo, Huang, Li, and Chang}]{luo2016text}
Yubo Luo, Yongfeng Huang, Fufang Li, and Chinchen Chang. 2016.
\newblock Text steganography based on ci-poetry generation using markov chain
  model.
\newblock \emph{KSII Transactions on Internet and Information Systems (TIIS)},
  10(9):4568--4584.

\bibitem[{Malik et~al.(2021)Malik, Sanjay, Guha, Hazarika, Nigam, Bhattacharya,
  and Modi}]{malik2021semantic}
Vijit Malik, Rishabh Sanjay, Shouvik~Kumar Guha, Angshuman Hazarika, Shubham
  Nigam, Arnab Bhattacharya, and Ashutosh Modi. 2021.
\newblock Semantic segmentation of legal documents via rhetorical roles.
\newblock \emph{arXiv preprint arXiv:2112.01836}.

\bibitem[{Modi et~al.(2023)Modi, Kalamkar, Karn, Tiwari, Joshi, Tanikella,
  Guha, Malhan, and Raghavan}]{legaleval-2023}
Ashutosh Modi, Prathamesh Kalamkar, Saurabh Karn, Aman Tiwari, Abhinav Joshi,
  Sai~Kiran Tanikella, Shouvik Guha, Sachin Malhan, and Vivek Raghavan. 2023.
\newblock {S}em{E}val-2023 {T}ask 6: {L}egal{E}val: {U}nderstanding {L}egal
  {T}exts.
\newblock In \emph{Proceedings of the 17th International Workshop on Semantic
  Evaluation (SemEval-2023)}, Toronto, Canada. Association for Computational
  Linguistics (ACL).

\bibitem[{Mohit(2014)}]{mohit2014named}
Behrang Mohit. 2014.
\newblock Named entity recognition.
\newblock \emph{Natural language processing of semitic languages}, pages
  221--245.

\bibitem[{Nadeau and Sekine(2007)}]{nadeau2007survey}
David Nadeau and Satoshi Sekine. 2007.
\newblock A survey of named entity recognition and classification.
\newblock \emph{Lingvisticae Investigationes}, 30(1):3--26.

\bibitem[{Paszke et~al.(2019)Paszke, Gross, Massa, Lerer, Bradbury, Chanan,
  Killeen, Lin, Gimelshein, Antiga et~al.}]{paszke2019pytorch}
Adam Paszke, Sam Gross, Francisco Massa, Adam Lerer, James Bradbury, Gregory
  Chanan, Trevor Killeen, Zeming Lin, Natalia Gimelshein, Luca Antiga, et~al.
  2019.
\newblock Pytorch: An imperative style, high-performance deep learning library.
\newblock \emph{Advances in neural information processing systems}, 32.

\bibitem[{Qiu et~al.(2020)Qiu, Sun, Xu, Shao, Dai, and Huang}]{qiu2020pre}
Xipeng Qiu, Tianxiang Sun, Yige Xu, Yunfan Shao, Ning Dai, and Xuanjing Huang.
  2020.
\newblock Pre-trained models for natural language processing: A survey.
\newblock \emph{Science China Technological Sciences}, 63(10):1872--1897.

\bibitem[{Ri et~al.(2021)Ri, Yamada, and Tsuruoka}]{ri2021mluke}
Ryokan Ri, Ikuya Yamada, and Yoshimasa Tsuruoka. 2021.
\newblock mluke: The power of entity representations in multilingual pretrained
  language models.
\newblock \emph{arXiv preprint arXiv:2110.08151}.

\bibitem[{Ri et~al.(2022)Ri, Yamada, and Tsuruoka}]{ri-etal-2022-mluke}
Ryokan Ri, Ikuya Yamada, and Yoshimasa Tsuruoka. 2022.
\newblock \href {https://aclanthology.org/2022.acl-long.505} {m{LUKE}: {T}he
  power of entity representations in multilingual pretrained language models}.
\newblock In \emph{Proceedings of the 60th Annual Meeting of the Association
  for Computational Linguistics (Volume 1: Long Papers)}. Association for
  Computational Linguistics.

\bibitem[{Shackelford et~al.(2016)Shackelford, Russell, and
  Kuehn}]{shackelford2016unpacking}
Scott~J Shackelford, Scott Russell, and Andreas Kuehn. 2016.
\newblock Unpacking the international law on cybersecurity due diligence:
  Lessons from the public and private sectors.
\newblock \emph{Chi. J. Int'l L.}, 17:1.

\bibitem[{Shan et~al.(2022)Shan, Baskaran, Yi, Ranek, Stanley, and
  Oliva}]{shan2022transparent}
Siyuan Shan, Vishal~Athreya Baskaran, Haidong Yi, Jolene Ranek, Natalie
  Stanley, and Junier~B Oliva. 2022.
\newblock Transparent single-cell set classification with kernel mean
  embeddings.
\newblock In \emph{Proceedings of the 13th ACM International Conference on
  Bioinformatics, Computational Biology and Health Informatics}, pages 1--10.

\bibitem[{Shan et~al.(2021)Shan, Li, and Oliva}]{shan2021nrtsi}
Siyuan Shan, Yang Li, and Junier~B Oliva. 2021.
\newblock Nrtsi: Non-recurrent time series imputation.
\newblock \emph{arXiv preprint arXiv:2102.03340}.

\bibitem[{Wang et~al.(2020)Wang, Hou, Che, and Liu}]{wang2020static}
Yuxuan Wang, Yutai Hou, Wanxiang Che, and Ting Liu. 2020.
\newblock From static to dynamic word representations: a survey.
\newblock \emph{International Journal of Machine Learning and Cybernetics},
  11:1611--1630.

\bibitem[{Xu et~al.(2018)Xu, Shan, Qiu, Jia, Shen, Wang, Shi, Eric, and
  Chang}]{xu2018end}
Yan Xu, Siyuan Shan, Ziming Qiu, Zhipeng Jia, Zhengyang Shen, Yipei Wang,
  Mengfei Shi, I~Eric, and Chao Chang. 2018.
\newblock End-to-end subtitle detection and recognition for videos in east
  asian languages via cnn ensemble.
\newblock \emph{Signal Processing: Image Communication}, 60:131--143.

\bibitem[{Yamada et~al.(2020)Yamada, Asai, Shindo, Takeda, and
  Matsumoto}]{yamada-etal-2020-luke}
Ikuya Yamada, Akari Asai, Hiroyuki Shindo, Hideaki Takeda, and Yuji Matsumoto.
  2020.
\newblock \href {https://doi.org/10.18653/v1/2020.emnlp-main.523} {{LUKE}: Deep
  contextualized entity representations with entity-aware self-attention}.
\newblock In \emph{Proceedings of the 2020 Conference on Empirical Methods in
  Natural Language Processing (EMNLP)}. Association for Computational
  Linguistics.

\bibitem[{Young(2009)}]{young2009legal}
Andy Young. 2009.
\newblock The legal duty of care for nurses and other health professionals.
\newblock \emph{Journal of Clinical Nursing}, 18(22):3071--3078.

\bibitem[{Zhang and Zhang()}]{zhang11fusion}
Jun Zhang and Xuan Zhang.
\newblock Fusion model based on attention mechanism to study energy-saving and
  emission-reduction schemes for sports competitions under the background of
  carbon neutrality.
\newblock \emph{Frontiers in Ecology and Evolution}, 11:1212732.

\end{thebibliography}
